Short Paper*

# Performance Evaluation of Regression Models in Predicting the Cost of Medical Insurance

Jonelle Angelo S. Cenita
Richwell Colleges, Incorporated, Philippines
cenita1994@gmail.com
(corresponding author)

Paul Richie F. Asuncion
Polytechnic College of Botolan, Philippines
paulasuncion@pcbzambales.com

Jayson M. Victoriano
Bulacan State University, Philippines
jayson.victoriano@bulsu.edu.ph

## Abstract

*Purpose* – The study aimed to evaluate the regression models' performance in predicting the cost of medical insurance. The Three (3) Regression Models in Machine Learning namely Linear Regression, Gradient Boosting, and Support Vector Machine were used. The performance will be evaluated using the metrics RMSE (Root Mean Square), r2 (R Square), and K-Fold Cross-validation. The study also sought to pinpoint the feature that would be most important in predicting the cost of medical insurance.

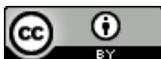

*Method* – The methodology of the study is anchored on the knowledge discovery in databases (KDD) process. (KDD) process refers to the overall process of discovering useful knowledge from data.

*Results* – The performance evaluation results reveal that among the three (3) Regression models, Gradient boosting received the highest r2 (R Square) 0.892 and the lowest RMSE (Root Mean Square) 1336.594. Furthermore, the 10-Fold Cross-validation weighted mean findings are not significantly different from the r2 (R Square) results of the three (3) regression models. In addition, Exploratory Data Analysis (EDA) using a box plot of descriptive statistics observed that in the charges and smoker features the median of one group lies outside of the box of the other group, so there is a difference between the two groups.

*Conclusion* – In conclusion, Gradient boosting appears to perform better among the three (3) regression models. K-Fold Cross-Validation concluded that the three (3) regression models are good. Moreover, Exploratory Data Analysis (EDA) using a box plot of descriptive statistics ceases that the highest charges are due to the smoker feature.

*Recommendations* – Gradient boosting model be used in predicting the cost of medical insurance.

*Research Implications* – Utilizing an accurate regression model to predict medical costs can aid medical insurance organizations in prioritizing the allocation of limited care management resources as it plays a role in the development of insurance policies.

## INTRODUCTION

In the study of Tkachenko et al. (2018), one of the key points of the development of the modern healthcare system is medical insurance. Also, the most crucial issue in this field is the prediction of individual medical insurance costs. A medical insurance business can only be profitable if it can collect greater money than it must pay for the beneficiaries' medical care. With that, it is necessary to pinpoint the feature that is most important in predicting the cost of medical insurance as it establishes the actuarial tables that adjust the price of yearly premiums higher or lower in accordance with the anticipated treatment costs. Accurate cost predictions can aid health insurers to choose insurance plans with appropriate deductibles and premiums and medical insurance organizations in prioritizing the allocation of limited care management resources as it plays a role in the development of insurance policies (Milovic & Milovic, 2012 ; Kumar et al., 2010).

According to Jordan & Mitchell (2015), machine-learning technology rapidly growing in technical fields because of the growing explosion in the availability of online data along with the advancement of computing technology and storage solutions. Panay et al. (2019) stated that researchers and practitioners have utilized a variety of machine-learning algorithms to analyze medical data to calculate medical insurance costs. Also, several machine-learning approaches have been applied to medical data analysis in studies. According to Muhammad & Yan (2015), there are three different types of machine learning can be distinguished: (1) supervised machine learning, which is task-driven and uses all labeled data for classification and regression; (2) unsupervised machine learning, which is data-driven and uses all unlabeled data for clustering; and (3) reinforcement learning, which uses mistakes as a decision making. This study uses supervised machine learning models to demonstrate and compare the accuracy of various regression models in predicting medical insurance costs.

In the big data era, the problem is deepened by the need for accurate and quick computations and the existence of a large number of data leads to the possibility of using a machine learning algorithm. Furthermore, the application of traditional regression approaches does not provide satisfactory results for the prediction of the medical insurance cost because of the big data problem (Mladenovic et al., 2020; Tkachenko et. al., 2018). According to Roopa and Asha (2019), regression models are used to predict and forecast the independent and dependent variables. Moreover, numerous types of regression models can be used and it is necessary to compare the various regression models to identify the most accurate for predicting the cost of medical insurance.

A statistical method known as regression analysis is used to establish the link between a single dependent (criterion) variable and one or more independent (predictor) variables. Following a linear combination of the predictors, the analysis produces a predicted value for the criterion (Palmer & O'Connel, 2009). Research by Marill (2004), claimed that linear regression is one of the common regression models to derive the regression line and is a popular technique because it can demonstrate mathematically and visually the relationship between important variables. The study by Maulud and Abdulazeez (2020), stated that linear regression is a mathematical approach that is used to perform predictive analysis, and its allowed continuous or mathematical variables projections. It is also a mathematical research method that possible to measure the predicted effects and model them against multiple input variables (Lim, 2019). Evaluation of the model is one of the crucial stages in machine learning studies. Comparing the trained model predictions with the actual (observed) data from the test data set is the goal of the evaluation (Botchkarev, 2018).

The study aimed to evaluate the regression models the performance in predicting the cost of medical insurance using the Kaggle dataset titled Medical Cost Personal

Datasets. The Three (3) Regression Models in Machine Learning namely Linear Regression, Gradient Boosting, and Support Vector Machine were used. The performance will be evaluated using the metrics RMSE (Root Mean Square) and r2 (R Square) that quantify how well the regression model fits a dataset then Ten (10) cross-validation will also be performed. The study also sought to pinpoint the feature that would be most important in predicting the cost of medical insurance.

## METHODOLOGY

The methodology of the study is anchored on the knowledge discovery in databases (KDD) process (Figure 1). (KDD) process refers to the overall process of discovering useful knowledge from data. In addition, the steps in the KDD Process are as followed, Selection, Pre-processing, Transformation, Data mining, and Interpretation/evaluation (Fayyad et al., 1996).

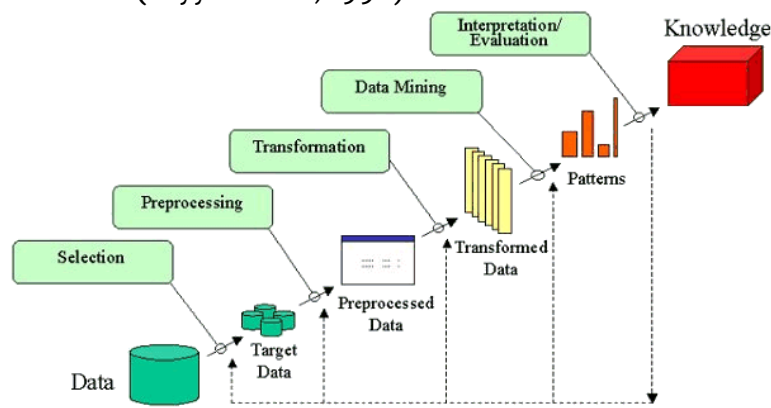

*Figure 1.* KDD Process

### Selection (Data Source)

The data sets originated from kaggle.com with Datasets titled Medical Cost Personal Datasets have four (4) numerical features (age, BMI, children, charges) with two (2) int and two (2) floats as datatype and two (3) categorical features (smoker, sex, and region) with three (3) objects as a datatype. The dataset has seven (7) features with non-null attributes and has a total of one thousand three hundred eighty-eight (1,388) entries in each column.

### Preprocessing (Data Analysis)

Figure 2 shows that there is an outlier between the charges feature to region, children, sex, and smoker features. Also, it reveals that the outlier begins in the charges cost amounting to seventeen thousand five hundred (17,500). The identified outliers that have charges value greater than seventeen thousand five hundred (17,500) are removed from the datasets.

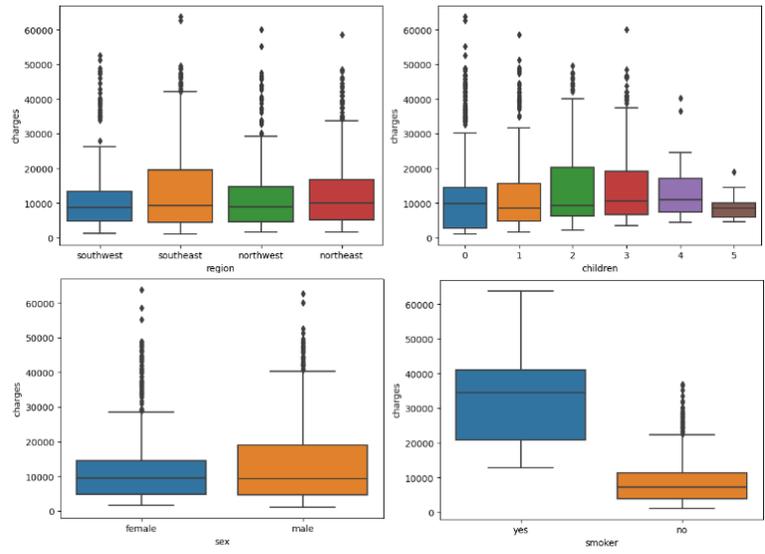

*Figure 2*. Exploratory Data Analysis (EDA) of Charges to region, children, sex, and smoker

## *Transformation (Data processing)*

The categorical features (smoker, sex, and region) with objects as datatype are converted into numerical features with int as data type (Figure 3). With that, the dataset has seven (7) features with non-null attributes and has a total of one thousand seventeen (1,017) in each column. Creating dependent as y and independent as x features, charges feature is the dependent feature while age, sex, bmi, children, smoker, and region features are the independent feature. Furthermore, The dataset is split into two training (80%) and test (20%), the training dataset undergoes fit_transform() while the test dataset will undergoes StandardScaler().

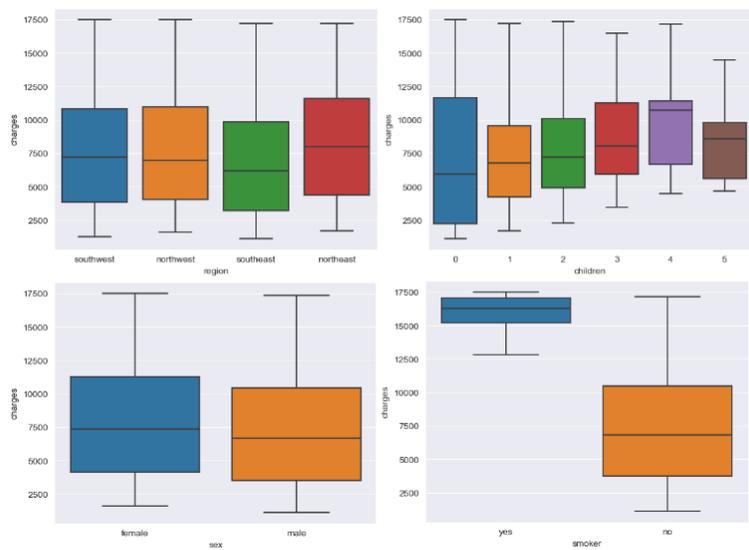

*Figure 3*. Exploratory Data Analysis (EDA) of Charges to region, children, sex, and smoker

Figure 3 shows that there is no outlier after the data processing. Furthermore, the median of one group lies within the box of the other groups which indicates that so there is no difference among the group. However, in the smoker feature, the median of one group lies outside of the box of the other group, so there is a difference between the two groups and it suggests that the highest charges are due to the smoker feature.

## Data Mining

### Regression models

According to Wu et al. (2019), Regression is a technique applied to two theories. The first is the theory that regression analyses are frequently employed for forecasting and prediction, with significant overlaps between their application and machine learning. The second theory is that regression analysis can occasionally be utilized to determine the relationships between the dependent and independent variables.

Regression models are used to predict and forecast the independent and dependent variables (Roopa & Asha, 2019). Also, in the book of Seber and Lee (2012), The goal of regression modeling is to create mathematical representations that characterize or explain potential relationships between variables.

### Linear regression

According to Acharya et al. (2019), linear regression is one of the simplest regression models for predicting outcomes (Figure 4). Also, the correlation between the independent and dependent variables is modeled. In a case model with a single independent variable, simple linear regression uses the formula of Linear Regression (Equation 1) to define the dependence of the variable (Maulud & Abdulazeez, 2020).

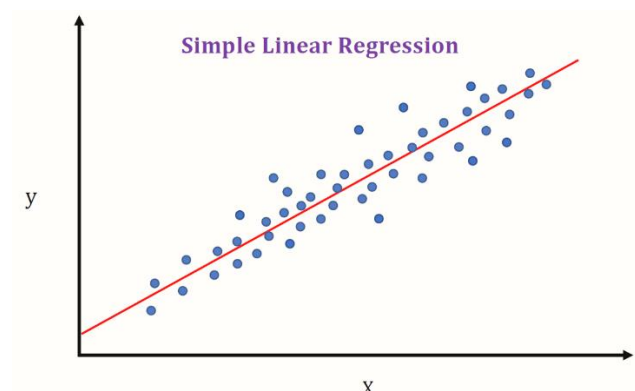

*Figure 4.* Linear Regression Model

$$y = β_0 + β_1 x + ε$$   *Equation 1. Linear Regression*

*Gradient boosting*

Gradient boosting is one of the techniques that allow for the recursive fitting of a weak learner to the residual, improving model performance with an ever-increasing number of iterations (Figure 5). It can automatically discover complex data structures, including nonlinearity and high-order interactions, even in the context of hundreds, thousands, or tens of thousands of potential predictors. The regression model would try to find a function that can accurately describe the data. However, the Gradient boosting function can only be an approximation of the data distribution and there must be errors:

$$y_i = F_1(X_i) + error_{1i}$$

*Equation 2. Gradient Boosting*

Where $X_i$ is a vector of predictors and $Y_i$ is the result variable. Assume that X and y relationship is not fully stated and that the $F_1(X_i)$ function is a weak learner. The error or residual in this instance has some correlation with y rather than being white noise (Zhang, Z., et al., 2019).

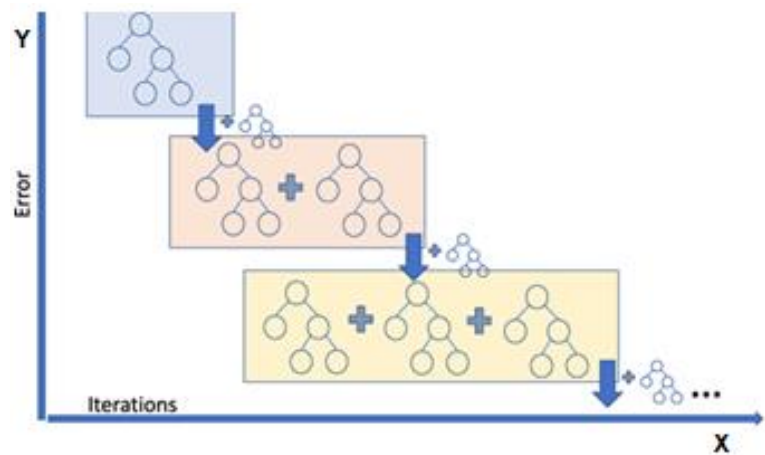

*Figure 5.* Gradient Boosting Model

*Support Vector Machine*

In the study of Maulud and Abdulazeez (2020), Support Vector Machines (SVM), sometimes known as SVR, SVR both linear and nonlinear regression. SVR seeks to fit as many instances on the street while reducing margin violations rather than aiming to fit the largest street between two classes while limiting margin violations and the hyperparameter epsilon controls the street's width (Figure 6).

According to Parbat and Chakraborty (2020), the generalized equation for hyperplane is represented in Equation 3, where w is weights and b is the intercept at X = 0. The margin of tolerance is represented by epsilon ε

$$y = wX + b$$

*Equation 3. Support Vector Machine*

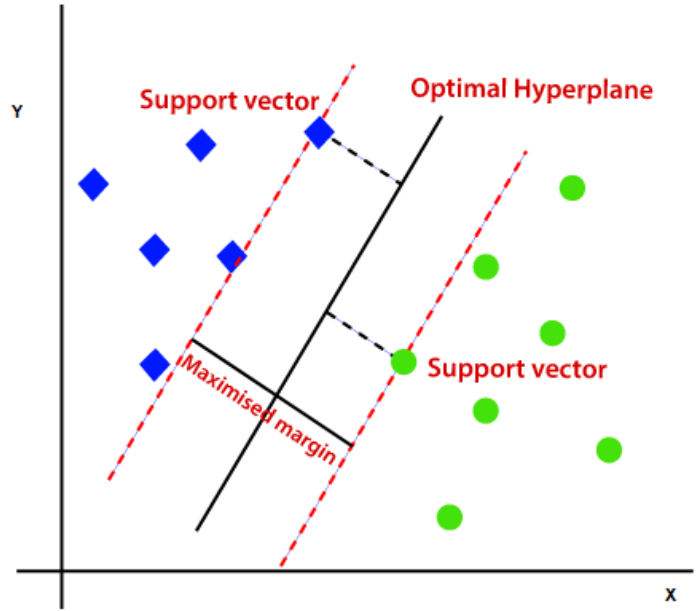

*Figure* 6. Support Vector Machine Model

**Performance Evaluation**

The factor that can be used to evaluate the performance of a regression model is defined in the work by Alexander et al. (2015).

**r2 (R Square)**

Equation 4 has been utilized in several scenarios in the literature in conjunction with training and test data when dealing with both linear and nonlinear regression models. Also, good models give small residuals

$$R2 = 1 - (RSS/TSS) \qquad \text{\textit{Equation 4.} \textit{R Squared}}$$

**RMSE (Root Mean Square)**

Equation 5 is an estimate of the standard deviation of residuals from the model.

$$RMSE = \sqrt{\Sigma(P_i - O_i)2 / n} \qquad \text{\textit{Equation 5.} \textit{RMSE (Root Mean Square)}}$$

**K-fold Cross-Validation**

Equation 6 is the most often used technique for estimating model predictability. Also, it is likely to give an overly optimistic assessment of the model's predictive power.

$$MSE = (1/n)*\Sigma(y_i - f(x_i))2 \qquad \text{\textit{Equation 6.} \textit{K-fold Cross-Validation}}$$

# RESULTS

## *Interpretation / Evaluation*

Three (3) regression models were used: Linear Regression, Gradient Boosting, and Support Vector Machine to evaluate the regression models' performance.

Also, the regression models undergo K-Fold Cross-Validation to evaluate the model performance and predict new data that hasn't been tested previously, which estimates the regression model's accuracy. Using the training dataset, it processed ten (10) consecutive times with different splits each time. The weighted mean of the K-Fold Cross-Validation will be computed.

Table 1. r2 and RMSE

| Regression Models | r2 (R Square) | RMSE (Root Mean Square) |
|---|---|---|
| Gradient boosting | 0.892 / 89% | 1336.594 |
| Linear regression | 0.881 / 88% | 1402.51 |
| Support vector machine | 0.881 / 88% | 1403.781 |

Table 1 shows that the Gradient boosting has the highest r2 value of 89%, indicating that the model can explain more variability of observed data. Furthermore, the r2 (R Square) also known as accuracy of Gradient boosting is 1% higher than the linear regression and the support vector machine. In light of this, the Gradient boosting model with 89% accuracy suggests that this model performed better than the rest.

Moreover, the Gradient boosting has the lowest RMSE value of 1336, which suggests that it is able to fit the dataset the best. Furthermore, the RMSE (Root Mean Square) also known as Residuals of Gradient boosting are 66 lower than linear regression and 67 lower than support vector machine.

Gradient boosting appears to perform better than linear regression and vector machines, according to results of performance evaluation of regression models.

Table 2. 10-Fold Cross-validation weighted mean

| Regression Models | K-Fold Cross-Validation |
|---|---|
| Gradient boosting | 0.879 / 88% |
| Linear regression | 0.860 / 86% |
| Support vector machine | 0.856 / 86% |

Table 2 shows that the three (3) regression models' 10-fold cross-validation weighted mean results are not significantly different from the r2 (R Square) also known as accuracy results of the three (3) regression models. This led to the conclusion that the three (3) regression models are good.

In addition, the Gradient boosting has the highest 10-fold cross-validation weighted mean of 88% followed by linear regression and support vector machine with 86%. This suggests that among the three (3) models gradient boosting model performed better.

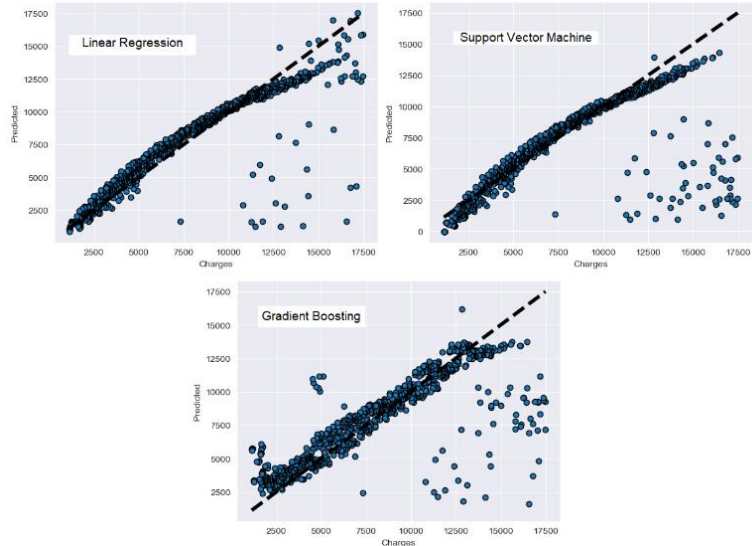

*Figure 7.* 10-Fold Cross-validation

The scatter plot in Figure 7 displays the 10-fold cross-validation results for the three regression models, and it demonstrates that the accuracy is inversely correlated with the residuals or prediction error.

Using descriptive statistics, a Box plot is a type of chart often used in explanatory data analysis and a Box plot visually shows the distribution of numerical data and skewness by displaying the data quartiles (or percentiles) and averages. The box plot in Figure 8 shows that there is a difference between the two groups by showing that the boxes or interquartile range do not overlap. Additionally, each group's middle line, or median, completely lies outside of each box, indicating that there is probably a difference between the two groups. The EDA contends that the smoking feature is to cause for the highest medical insurance.

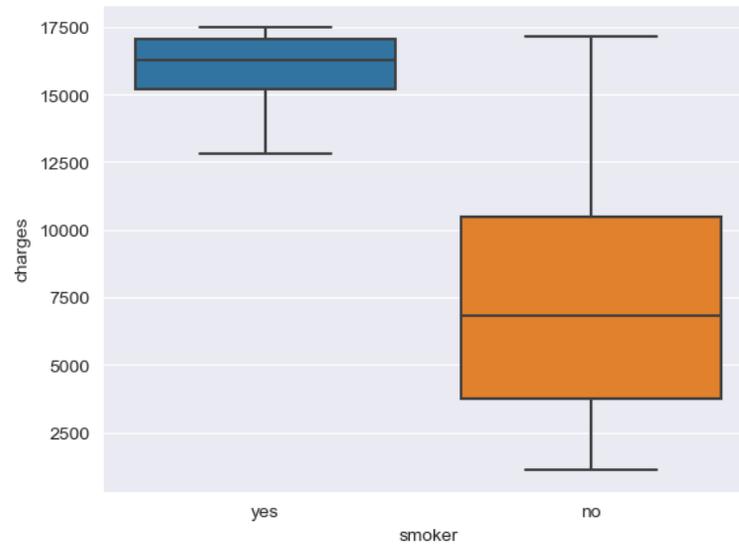

*Figure 8*. Exploratory Data Analysis (EDA) of Charges and smoker

## DISCUSSION

Many researchers utilize machine learning algorithms because they offer effective outcomes. The exploratory data analysis using a box plot of descriptive statistics reveals that the smoker factor is the greatest contributor to the increase in chargers because the median of one group lies outside of the box of the other group, which indicates there is a difference between the two groups. Considering this, it is strongly recommended that the highly contributory factor to the cost of medical insurance is smoking.

The three (3) Machine Learning regression models – Linear Regression, Gradient Boosting, and Support Vector Machine, were applied in this study. The experiment shows that the Gradient boosting with 89% r2 (R Square) produced the highest accuracy for the prediction of the medical insurance cost and proved that the model can explain more variation in the observed data. Furthermore, it has the lowest RMSE (Root Mean Squared) of 1336 residuals and it suggests that it is the best fit for the dataset. To validate the results of the performance evaluation of the regression models, it undergoes 10-fold cross-validation and the results are not significantly different from the r2 (R Square) accuracy results of the three (3) regression models.

## CONCLUSIONS AND RECOMMENDATIONS

In conclusion, Gradient boosting appears to perform better among the Three (3) Regression Models in Machine Learning namely Linear Regression, Gradient Boosting, and Support Vector Machine. K-Fold Cross-Validation concluded that the three (3) regression models are good. Moreover, Exploratory Data Analysis (EDA) ceases that the highest charges are due to the smoker feature. Therefore, it is favorable and feasible to suggest that the Gradient boosting model, which outperforms the other two regression models in terms of accuracy in predicting the cost of medical insurance, be used.

## IMPLICATIONS

Utilizing an accurate regression model to predict medical costs can aid medical insurance organizations in prioritizing the allocation of limited care management resources as it plays a role in the development of insurance policies.

## ACKNOWLEDGEMENT

The researchers want to express deep gratitude to the ICpEP and IRCCETE 2023. Sincere appreciation to the conference organizers for organizing this event and for including this paper in the conference.

The researchers would like to acknowledge and express gratitude to the organizations with which we are affiliated, including Bulacan State University, Polytecnic College of Botolan, and Richwell Colleges, Inc. The researchers are grateful for your kind encouragement and motivation to carry out this research. Also, for the Kaggle.com for the dataset.

## DECLARATIONS

### *Conflict of Interest*

The authors declare no conflict of interest.

### *Informed Consent*

Not applicable, datasets came from Kaggle.com a subsidiary of Google that allows users to find datasets they want to use in building AI models and publish datasets.

### *Ethics Approval*

Not applicable, the dataset was not collected by the authors, and it is publicly available at Kaggle.com.

**Author's Biography**

Jonelle Angelo S. Cenita is a college instructor at Richwell Colleges, Incorporated. He is currently enrolled in the program of Doctor of Information Technology at La Consolacion University Philippines and is a graduate of Master of Science in Information technology and Bachelor of Science in Information Technology.

Paul Richie F. Asuncion is a faculty member of Polytechnic College of Botolan in Botolan, Zambales, and is currently designated as the dean of the Institute of Computing Studies. He is currently pursuing Doctor in Information Technology at La Consolacion University Philippines in Malolos, Bulacan. He obtained his master's degree at the University of the Philippines Open University under the Master of Information Systems program. His research interest includes Information System development, data mining, artificial intelligence, and things relative to IT education administration.

Dr. Jayson M Victoriano is the Program Chair of BS Data Science at Bulacan State University-Sarmiento Campus, He is also a member of the prestigious National Research Council of the Philippines, He is also the Current Director for Research and Innovation for External Campuses at the same University.